# From Texts to Structured Documents: The Case of Health Practice Guidelines

Amanda Bouffier

Laboratoire d'Informatique de Paris-Nord
Université Paris 13 and CNRS UMR 7030
99, av. J.-B. Clément – F-93430 Villetaneuse

firstname.lastname@lipn.univ-paris13.fr

**Abstract.** This paper describes a system capable of semi-automatically filling an XML template from free texts in the clinical domain (practice guidelines). The XML template includes semantic information not explicitly encoded in the text (pairs of conditions and actions/recommendations). Therefore, there is a need to compute the exact scope of conditions over text sequences expressing the required actions. We present in this paper the rules developed for this task. We show that the system yields good performance when applied to the analysis of French practice guidelines.

## 1 Introduction

During the past years, clinical practices have considerably evolved towards standardization and effectiveness. A major improvement is the development of practice guidelines [1]. However, even if widely distributed to hospitals, doctors and other medical staff, clinical practice guidelines are not routinely fully exploited. There is now a general tendency to transfer these guidelines to electronic devices (via an appropriate XML format). This transfer is justified by the assumption that electronic documents are easier to browse than paper documents.
The Guideline Elements Model (GEM) is an XML-based guideline document model that can store and organize the heterogeneous information contained in practice guidelines [2]. It is intended to facilitate translation of natural language guideline documents into a format that can be processed by computers. The main element of GEM, `knowledge component`, contains the most useful information, especially sequences of conditions and recommendations. Our aim is to format these documents, which have been written manually without any precise model, according to the GEM DTD.
One of the main problem for the task is that the scope of the conditional segments (i.e all the recommendation segments that have to be linked with a condition) may exceed the sentence boundaries and, thus, include several sentences. In other words, sequences of conditions and recommendations correspond to *discourse* structures.
Discourse processing requires the recognition of heterogeneous linguistic features (especially, the granularity of relevant features may vary according to text genre [9]).
Following these observations, we made a study based on a representative corpus and automatic text mining techniques, in order to semi-automatically discover relevant linguistic features for the task and infer the rules necessary to accurately structure the practice guidelines.

The paper is organized as follow: first, we present the task and some previous approaches (section 2). We then describe the rules for text structuring (section 3) and the method used to infer them. We finish with the presentation of some results (section 4), before the conclusion.

## 2    Document Restructuring: the Case of Practice Guidelines

As we have previously seen, practice guidelines are not routinely fully exploited. To overcome this problem, national health agencies try to promote the electronic distribution of these guidelines.

### 2.1    Previous Work

Several attempts have already been made to improve the use of practice guidelines. For example, knowledge-based diagnostic aids can be derived from them [3]. GEM is an intermediate document model, between pure text (paper practice guidelines) and knowledge-based models like GLIF [4]. GEM is thus an elegant solution, independent from any theory or formalisms, but compliant with other frameworks. Previous attempts to automate the translation process between the text and GEM are based on the analysis of isolated sentences and do not compute the exact scope of conditional segments [5].

### 2.2    Our Approach

Our aim is to semi-automatically fill a GEM template from existing guidelines: the algorithm is fully automatic but the result needs to be validated by experts to yield adequate accuracy. We first focus on the most important part of the GEM DTD `knowledge Component` which is sequences of conditions and recommendations.
We propose a two-step strategy: 1) basic segments (conditions and recommendations) are recognized and 2) the scope of the conditional segments is computed. In this paper, we focus on the second step, which is the most difficult one and has not been solved by previous systems. What is obtained in the end is a tree where the leaves are recommendations and the branching nodes are conditional segments, as shown on Figure 1. All the children of a node are under the scope of the parent node.

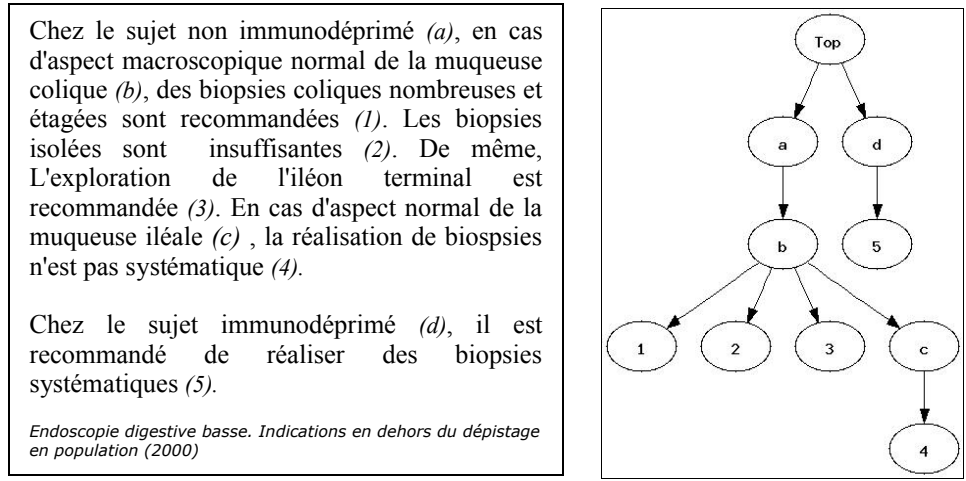

Figure 1: From text to a tree representing the scope of conditional segments

## 3 Structuring rules

We set up a representative corpus in order to infer a set of rules able to decide if a segment *s* is (or is not) under the scope of a conditional segment *c*.

### 3.1 Material & method

The training corpus consists of 25 French Practice Guidelines (about 150 000 words, see http://anaes.fr). This corpus has been annotated by a domain expert, who had to manually recognize conditions and recommendations, and link them according to the tree structure described in 2.2. We have built on this basis a set of examples, each example being a couple (*c, s*) where *c* is a conditional segment and *s* is a segment under the scope of *c* (positive examples) or not (negative examples). All the examples are represented by a set of 17 potential interesting linguistic features. The list of these features has been identified from the relevant literature (e.g. [6] [7] [8]) and a manual study of the practice guidelines. More precisely, the features belong to the 4 following categories:

**1) Material text structure**. Let's take the feature: "has_the_same_visual_position*(c,s)*". Its value for the couple *(a,d)* in Figure 1 is "true" because *(a)* and *(d)* are both at the beginning of a paragraph. Another attribute concerns the location of the condition: "is detached*(c)*". The value of this feature for *(c)* in the figure 1 is "detached" because *(c)* is located at the sentence-initial position.

**2) Lexical relations**. Let's take the feature "have_terms_with_an_antonymic_relation(c,s)". Its value is "true" for the couple *(a,d)* because the terms *immunodéprimé* and *non immunodéprimé* are antonyms.

**3) Discourse connectors**. Let's take the feature "begins_with_a_coordination_ marker *(s)*". Its value is "true" for *(a,3)* because the proposition begins with *De même (also...)*.

**4) Co-reference relations.** This feature is useful to recognize co-references. In the medical domain, demonstrative noun phrases ("dans <u>ce cas</u>", "in <u>this case</u>...".) is the most common way of marking co-reference.

We then used data mining algorithms (in particularly attributes selectors and rules learning algorithms) in order to statistically validate the potential interest of the different features, understand their relative contribution and derive the set of structuring rules.

### 3.2 The rules organized in knowledge levels

One of the main result of our study is that features related to the material structure of the text are the most discriminating ones for the task, using an "Information Gain" measure. First, the location of the condition in the sentence (detached or not) is especially important. If it is detached, it exerts in 70% of the cases an influence downward from the sentence in which it is located. Conversely, if the condition is integrated inside the sentence, its scope is limited to the sentence boundaries in 80 % of the cases. More generally, the scope of a conditional segment often complies with the material text structure. Thus, the rules which involve these features correspond to **norms**. This can be explained by the style of writing used for the "guidelines" text genre, that often makes use of visual information. The other kinds of attributes (discourse connectors, lexical relations co-coreference informations) are less discriminating simply because they are less frequent. Nevertheless, they are sometimes important since they may violate a norm and suggest a more accurate way of structuring the document. The rules which involve these features are called **exceptions**. Therefore, we have organized the structuring rules according the ability of a feature to contribute to the solution (norms vs. exceptions).

The **norms** represent the most discriminating rules and involve the most salient features which belong to the category of the material text structure.

- By default, IF c is syntactically integrated AND *s* is in another sentence of *c* THEN *s* is excluded from the scope of *c*. Conversely, IF *c* is syntactically integrated AND *s* is in the sentence of *c* THEN *s* is under the scope of *c*.
- By default, IF *c* is detached from the sentence AND *s* is in the same position in the material structure than *c* THEN *s* is excluded from the scope of *c*. Conversely, IF *s* is in the same position in the material structure than *c* AND *s* and *c* are in the same paragraph THEN *s* is included in the scope of *c*.

These two rules can be violated if another set of rules called **exceptions** suggest a more accurate way of structuring the text. A first set of rules suggest an inclusion of *s* under the scope of *c*. For example, the following rule belongs to this category:

- IF s is in a co-reference relation with *c* THEN *s* is under the scope of *c*.

Conversely, some other rules suggest an exclusion of *s*. For example:
- IF *s* is preceded by a discourse coordination connector THEN *s* is excluded from the scope of *c*.

## 4 Evaluation

We evaluated the approach on a corpus that has not been used for training. The evaluation of basic segmentation gives the following results: .92 P&R[1] for conditional segments and .97 for recommendation segments. The scope of conditions is recognized with accuracy above .7. This result is encouraging, especially considering the large number of parameters involved in discourse processing. In most of successful cases the scope of a condition is recognized by the default rule (default segmentation, see section 3).

## 5 Conclusion

We have presented in this paper a system capable of performing automatic segmentation of clinical practice guidelines. Our aim was to automatically fill an XML DTD from textual input. The system is able to process complex discourse structures and to compute the scope of conditional segments spanning several propositions or sentences. Moreover, our system is the first one capable of resolving the scope of conditions over several recommendations.

---

[1] P&R is the harmonic mean of precision and recall (P&R = (2*P*R) / (P+R), corresponding to a F-measure with a $\beta$ factor equal to 1